\documentclass[runningheads]{llncs}
\usepackage{graphicx}
\usepackage{comment}
\usepackage{amsmath,amssymb} 
\usepackage{color}
\usepackage{multirow}
\usepackage{subfigure}
\usepackage{float}

\usepackage{microtype}

\usepackage[width=122mm,left=12mm,paperwidth=146mm,height=193mm,top=12mm,paperheight=217mm]{geometry}

\newcommand{\NUMSCANNETLAYOUT}{1151}

\begin{document}

\pagestyle{headings}
\mainmatter

\newcommand{\name}{Scene CAD}

\title{ \vspace{-0.5cm}SceneCAD: Predicting Object Alignments \\ and Layouts in RGB-D Scans \vspace{-0.5cm}} 

\titlerunning{SceneCAD}
%

\author{Armen Avetisyan$^{1}$ \and
Tatiana Khanova$^{3}$ \and
Christopher Choy$^{2}$ \and \\
Denver Dash$^{3}$ \and 
Angela Dai$^{1}$ \and
Matthias Nie{\ss}ner$^{1}$ }
\authorrunning{Avetisyan et al.}
%
\institute{Technical University of Munich \and Stanford University \and Occipital Inc.}


\maketitle

\begin{figure}[th!]
   \vspace{-0.5cm}
     \centering
        \includegraphics[width=\linewidth]{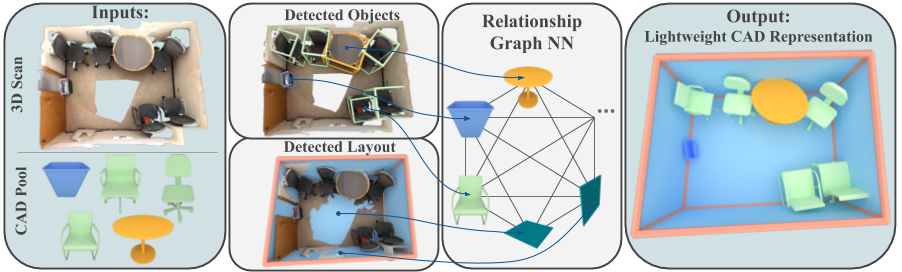}
   \label{fig:teaser}
   \vspace{-0.3cm}
       \caption{Our method takes as input a 3D scan and a set of CAD models. 
       We jointly detect objects and layout elements in the scene.
       For the objects, we find similar CAD models from the candidate set.
       Each detected object or layout component then forms a node in a graph neural network which estimates object-object relationships and object-layout relationships.
       This holistic understanding of the scene enables robust retrieval and alignment of CAD models to the scan as well as layout generation, resulting in a lightweight CAD-based representation of the scene.
       }

\end{figure}
       \vspace{-1cm}

\begin{abstract}
We present a novel approach to reconstructing lightweight, CAD-based representations of scanned 3D environments from commodity RGB-D sensors.
Our key idea is to jointly optimize for both CAD model alignments as well as layout estimations of the scanned scene, explicitly modeling inter-relationships between objects-to-objects and objects-to-layout.
Since object arrangement and scene layout are intrinsically coupled, we show that treating the problem jointly significantly helps to produce globally-consistent representations of a scene.
Object CAD models are aligned to the scene by establishing dense correspondences between geometry, and we introduce a hierarchical layout prediction approach to estimate layout planes from corners and edges of the scene. 
To this end, we propose a message-passing graph neural network to model the inter-relationships between objects and layout, guiding generation of a globally object alignment in a scene.
By considering the global scene layout, we achieve significantly improved CAD alignments compared to state-of-the-art methods, improving from 41.83\% to 58.41\% alignment accuracy on SUNCG and from 50.05\% to 61.24\% on ScanNet, respectively.
The resulting CAD-based representations makes our method well-suited for applications in content creation such as augmented- or virtual reality. 
\end{abstract}

\section{Introduction}

The recent progress of 3D reconstruction of real-world environments from commodity range sensors has spurred interest in using such captured 3D data for applications across many fields, such as content creation, mixed reality, or robotics.
State-of-the-art 3D reconstruction approaches can now produce impressively-robust camera tracking and surface reconstruction~\cite{newcombe2011kinectfusion,niessner2013hashing,choi2015robust,dai2017bundlefusion}.

Unfortunately, the resulting 3D reconstructions are not well-suited for direct use with many applications, as the geometric reconstructions remain incomplete (e.g., due to occlusions and sensor limitations), are often noisy or oversmoothed, and often consume a large memory footprint due to high density of triangles or points used to represent a surface at high resolution.
There still remains a notable gap between these reconstructions and artist-modeled 3D content, which are clean, complete, and lightweight~\cite{Gupta2015Aligning3M}.

Inspired by these attributes of artist-created 3D content, we aim to construct a CAD-based scene representation of an input RGB-D scan, with objects represented by individual CAD models and scene layout represented by lightweight meshes.
In contrast to previous approaches which have individually tackled the tasks of CAD model alignment~\cite{li2015database,avetisyan2019scan2cad,avetisyan2019end} and of layout estimation~\cite{murali2017indoor,liu2018floornet,chen2019floor}, we observe that object arrangement is typically tightly correlated with the scene layout.
We thus propose to jointly optimize for CAD model alignment and scene layout to produce a globally-consistent CAD-based representation of the scene.

From an input RGB-D scan along with a CAD model pool, we align CAD models to the scanned scene by establishing dense correspondences. 
To estimate the scene layout, we characterize the layout into planar elements, and propose a hierarchical layout prediction by first detecting corner locations, then predicting scene edges, and from sets of edges potentially presenting a layout plane, predicting the final layout.
We then propose a graph neural network architecture for optimizing the relationships between objects and layout by predicting object-object relative poses as well as object-layout support relationships.
This optimization guides both object and layout arrangement to be consistent with each other.
Our approach is fully-convolutional and trained end-to-end, generating a CAD-based scene representation of a scan in a single forward pass.

\medskip

\noindent
In summary, we present the following contributions:
\begin{itemize}
    \item We formulate a lightweight heuristic-free 3D layout prediction algorithm that hierarchically predicts corners, edges and then planes in an end-to-end fashion consisting of only $\approx 1M$ trainable parameters generating satisfactory layouts without the need for extensive heuristics.
    \item We present a scene graph network that learns relationships between objects and scene layout, enabling globally consistent CAD model alignments and results in a significant increase in prediction performance in both synthetic as well as real-world datasets. 
    \item We introduce a new richly-annotated real-world scene layout dataset consisting of \NUMSCANNETLAYOUT{} CAD shells and wireframes on top of the ScanNet RGB-D dataset, allowing large-scale data-driven training for layout estimation. 
\end{itemize}

\section{Related Work}

\paragraph{CAD model alignment}
%
Aligning an expert-generated 3D model or a 3D template to 3D scan data has been studied widely due to its wide range of applications, for instance motion capture~\cite{bogo2016keep}, 3D object detection and localization~\cite{drost20123d,engelmann2016joint,zakharov2019dpod}, and scene registration~\cite{Wald_2019_ICCV}.
Our aim is to leverage large-scale datasets of CAD models to reconstruct a lightweight, semantically-informed, high-quality CAD representation of an RGB-D scan of a scene.
Several approaches have been developed to retrieve and align CAD models from a shape database and align them in real time to a scan during the 3D scanning process~\cite{kim2013guided,li2015database}, although their use of handcrafted features for geometric scan-to-CAD matching limit robustness.

Zeng et al.~\cite{zeng20173dmatch} developed a learned feature extractor using a siamese network design for geometric feature matching, which can be employed for scan-to-CAD feature matching, though this remains difficult due to the domain gap between synthetic CAD models and real-world scans.
Avetisyan et al.~\cite{avetisyan2019scan2cad} proposed a scan-to-CAD retrieval and alignment approach leveraging learned features to detect objects in a 3D scan and establish correspondences across the domain gap of scan and CAD. 
They later built upon this work to develop a fully end-to-end trainable approach for this CAD alignment task~\cite{avetisyan2019end}.
For such approaches, each object is considered independently, whereas our approach exploits contextual information from object-object and object-layout to produce globally consistent CAD model alignment and layout estimation.

Other approaches retrieve and align CAD models to RGB images~\cite{lim2013parsing,Xiang2016ObjectNet3DAL,Sun_2018}; our work instead focuses on geometric alignment of CAD models and layout.

\paragraph{Graph neural networks and relational inference in 3D.}
Recent developments in graph inference and graph neural networks have shown significant promise for inference on 3D data. 
Recently, various approaches have viewed 3D meshes as graphs in order find correspondences between 3D shapes~\cite{bronstein2017geometric}, deform a template mesh to fit an image observation of a shape~\cite{wang2018pixel2mesh}, or generate a mesh model of an object~\cite{dai2019scan2mesh}, among other applications.
Learning on graphs has also shown promise for estimating higher-level relational information in scenes, as a scene graph.
3D-RelNet~\cite{kulkarni20193drelnet} predicts 3D shapes and poses from single RGB images and establish pairwise pose constraints between objects to improve overall prediction quality. 
Our approach is similarly inspired to establish relationships between objects; we additionally employ relationships between objects and structural components (i.e., walls, floors, and ceilings), which considerably inform object arrangement.
Armeni et al.~\cite{armeni20193d} propose a unified hierarchical structure that hosts building, room, and object relationships into one 3D scene graph. 
They leverage this graph structure to generate scene graphs from 2D images.
Our approach focuses on leveraging relational information to reconstruct imperfect scans with a CAD-based representation for each object and layout element.

\paragraph{Layout estimation.}
Various layout estimation approaches have been developed to infer structural information from RGB and RGB-D data.
Scan2BIM~\cite{murali2017indoor} generates building information models (BIM) from 3D scans by detecting planes and finding plausible intersections to produce room-level segmentation of floors, ceilings and walls under Manhattan-style constraints.
PlaneRCNN~\cite{liu2018planercnn} and PlaneNet~\cite{Liu_2018} propose deep neural network architectures to detect planes from RGB images and estimate their 3D parameters.
FloorNet~\cite{liu2018floornet} estimates a 2D Manhattan-style floorplan representation for an input RGB-D scan using a point-based neural network architecture.
Floor-SP~\cite{chen2019floor} relaxes the Manhattan constraints with an integer programming formulation, and produces more robust floorplan estimation. 
In contrast to these layout estimation approaches, our focus lies in leveraging global scene relations between objects as well as structural elements in order to produce a CAD-based representation of the scene.

\begin{figure*}[tp]
\begin{center}
   \includegraphics[width=\linewidth]{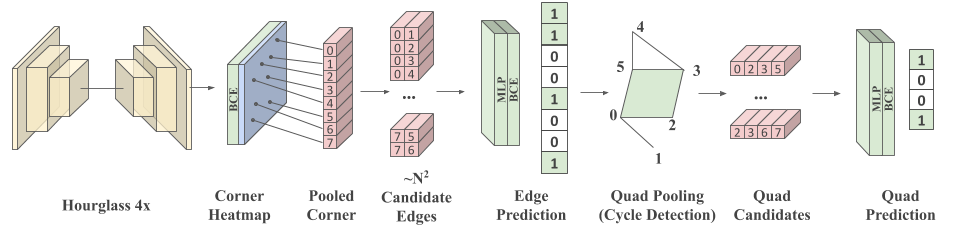}
    \end{center}
   \caption{Layout estimation as planar quad structures.
   Layout components are characterized as planar elements which are detected hierarchically. 
   From an input scene, corners of these layout elements are predicted in heatmap fashion leveraging non-maximum suppression. 
   From these predicted corners, edges are then predicted for each possible pair of corners as a binary classification task.
   From the predicted edge candidates, valid quads of four connected edges are considered as candidate layout elements, with a binary classification used to produce the final layout prediction.
   }
\label{fig:architecture}
\end{figure*}

\section{SceneCAD: Joint Object Alignment and Layout Estimation}
From an RGB-D scan and a CAD model dataset, our method jointly predicts CAD model alignment and layout estimation to produce a lightweight, CAD-based representation of the scene (Figure~\ref{fig:teaser}).
The input scan is represented as a sparse 3D voxel grid of the occupied surface geometry carrying fused RGB data.
The scan is first encoded by a series of sparse 3D convolutional layers~\cite{choy20194d} to produce a feature volume $F'$. The sparse output $F'$ is then densified into a dense 3D feature grid $F \in \mathbb{R}^{N_f \times N_x \times N_y \times N_z}$ where $N_F$ is the number of channels in the feature and $N_x, N_y$, and $N_z$ are the resolution of the feature along x, y, and z axis respectively. Note that the encoder serves as backbone for proceeding modules. Hence, $F$ is the input to the CAD alignment module as well as the layout estimation module.

Based on $F$, we detect objects along with their bounding box in the object detection module and layout planes in the layout detection module. 
We then establish our relational inference by formulating a message-passing graph neural network on the predicted objects and layout planes, where each node represents an object or layout plane, with losses on edge relationships representing relative poses and support. Finally, we predict a set of retrieved CAD models along with their $9$-DoF poses (3 translation, 3 rotation, and 3 scale) for every detected object.

The message-passing graph neural network helps to inform objects of both relations between other objects as well as with the scene layout, e.g., certain types of furniture such as beds and chairs are typically directly supported by a floor, chairs near a table often face the table.
This joint optimization thus helps to enable globally consistent CAD model alignment in the final output. 

\subsection{Layout Prediction}
The indoor scene of interest in our problem consists of planar or quadrilateral components such as walls, floors, and ceilings. However, some of these planar elements create complex geometry such as bars, beams, or other structures that effectively make template-matching approach to find the room layout challenging. Thus, we propose a bottom-up approach that predicts corners, edges, and planar elements sequentially to predict the room layout.
Our layout prediction pipeline is structured hierarchically: first predicting the corner locations, then predicting edges between the corners, and finally extracting quads from the predicted edges. We visualize the overview of the pipeline on Figure~\ref{fig:architecture}.

Corners are predicted by a convolutional network that decodes $F$ to its original dimension by predicting a heatmap; i.e. a voxel-wise score that indicates a \textit{cornerness} likeliness. The loss for this predicted heatmap is a voxel-wise binary cross-entropy classification loss in conjunction with a softmax and a negative log-likelihood over the entire voxel grid where the problem is formulated as a spatial multi-class problem. 
This is structured as an encoder-decoder, where the bottleneck lies at a spatial reduction of $4\times$.
Note that we make predictions for corners which have not been observed in the input scan (e.g., due to occlusions, c.f.). See supplemental material for a visual illustration of the layout prediction pipeline.  
From the output corner heatmap, we apply a non-maximum suppression to filter out weak responses, and define the final corner predictions as a set of xyz coordinates $\mathcal{V} = \{\mathbf{v}_i\}_i$, $\mathbf{v}_i = [x_i, y_i, z_i]$.

We the predict the layout edges from the predicted corners $\mathcal{V}$.
We construct the candidate set of edges by taking all pair-wise combinations of corners $\mathbf{e}_{ij} = (\mathbf{v}_i, \mathbf{v}_j)$ for all $i \in [1, ..., |\mathcal{V}|]$ and $j \in [1, ..., i - 1]$. We denote all edges as $\mathcal{E} = \{\mathbf{e}_{ij}\}_{ij}$.
From the pool of candidate edges we predict a set of edges that belongs to the scene structure using a graph neural network. 
Specifically, for each potential edge $\mathbf{e}_{ij} = (\mathbf{v}_i, \mathbf{v}_j)$, we extract corresponding features from the vertex prediction convolutional network, $F[\mathbf{v}_i], F[\mathbf{v}_j]$ where $F[\cdot]$ denotes the feature vector at the specified $x, y, z$ coordinate. We concatenate these features along with the normalized coordinates to form an input feature vector for each edge $\mathbf{f}_{\mathbf{e}_{ij}} = [F[\mathbf{v}_i], F[\mathbf{v}_j], \mathbf{N}(\mathbf{v}_i), \mathbf{N}(\mathbf{v}_j)]$.
For each edge we construct two feature descriptors with alternating order of corner features $\mathbf{f}_{\mathbf{e}_{ji}}$ to mitigate the effect of order dependency.
We feed these concatenated features into a graph network, which we train with edge-wise binary cross entropy loss against ground truth edges.
As the vertex predictions have uncertainty, we label edges with predicted vertices within a certain radius from the ground truth layout vertices to be positives.
This edge prediction limits the set of candidate layout quads which would otherwise be $O\left(\binom{|\mathcal{V}|}{4}\right)$.

From these predicted edges, we then compute the set of candidate layout quads as the set of planar, valid 4-cycles within these edges $\mathbf{q}_{ijkl} = \{\mathbf{e}_{ij}, \mathbf{e}_{jk}, \mathbf{e}_{kl}, \mathbf{e}_{li}\}$.
To detect valid cycles, we use the depth-first-search cycle detection algorithm
We predict the final set of layout quads as either positive or negative where the positive predictions constitute the scene layout, decomposed as quads.
The feature descriptor for a candidate quad is constructed by concatenating the features from $F$ corresponding to the corner locations of its vertices and normalized corner locations, $\mathbf{q}_{ijkl} = [F[\mathbf{v}_i], F[\mathbf{v}_j], F[\mathbf{v}_k], F[\mathbf{v}_l], \mathbf{N}(\mathbf{v}_i), \mathbf{N}(\mathbf{v}_j), \mathbf{N}(\mathbf{v}_k), \mathbf{N}(\mathbf{v}_l)]$. Similar to the edge features, every quad feature descriptor is 4-way permuted $\mathbf{q}_{jkli}, \mathbf{q}_{klij}$, and $\mathbf{q}_{lijk}$ in order to mitigate order-dependency.
This feature is input to an MLP followed by a binary cross entropy loss. From these predicted quads, we recover the scene layouts without heuristic post-processing.

\subsection{CAD Model Alignment}

Along with the room layout, we aim to find and align light-weight CAD models to objects in the scanned scene. To this end, we propose a CAD model alignment pipeline that detects objects, retrieves CAD models, and finds transformations that aligns the CAD model to the scanned scene.
First, we use a single-shot anchor-based object detector to identify objects~\cite{hou20193dsis}, using the features from the backbone we extracted ($\mathbf{F}$) from the previous stage. 
We then filter the predicted anchors with non-maximum suppression following the standard single-shot object detection pipeline~\cite{liu2016ssd}. 
Given this set of object bounding boxes $\mathcal{B}$, we extract $N_d \times N_d \times N_d$ feature volume $F_o$ for all $o \in [1,..., |\mathcal{B}|]$ from the feature map $F$ around the object anchor $a_o$. 
We use this feature volume for CAD model retrieval and alignment. 
A corresponding CAD model is retrieved by calculating an object descriptor of length 512 and searching the nearest neighbor CAD model from an shared embedding space. 
This shared embedding space is established by minimizing the distance between descriptors of scanned objects and their CAD counterpart with an L1 loss during training.  

Finally, given the nearest CAD model for all object anchors, we find dense correspondences between the CAD model and the feature volume $F_o$. Dense correspondences are trained through an explicit voxel-wise L1 regression loss. We use Procrutes~\cite{goodall1991procrustes} to estimate a rotation matrix and an L1 distance loss with respect to the groundtruth rotation matrix to further enhance correspondence quality. 
Note that the Procrutes method yields a transformation matrix through the Singular Value Decomposition which is differentiable, allowing for end-to-end training. 

\subsection{Learning Object and Layout Relationships}
From our layout prediction and CAD model alignment, we obtain a set of layout quads and aligned CAD models, both obtained independently from the same backbone features.
However, this can result in globally inconsistent arrangements; for instance, objects passing through the ground floor, or shelves misaligned with walls.
We thus propose to learn the object-layout as well as object-object relationships as a proxy loss used to guide the CAD model alignments and layout quads into a globally consistent arrangement.

We construct this relationship learning as a graph problem, where the set of objects and layout quads form the nodes of the graph. 
Edges are constructed between every object-object node-pair and every object-quad node-pair, forming a graph on which we formulate a message-passing graph neural network.

Each node of the graph is characterized by a feature vector of length 128.
For objects this feature vector is obtained by pooling the object feature volume to $8^3$ resolution, followed by linearization.
For layout quads, this feature vector is constructed by concatenating the features from $F$ or the associated corner locations, upon which an MLP is applied to obtain a $128$-dimensional vector.

\begin{figure}[tbh]
\begin{center}
   \includegraphics[width=0.9\linewidth]{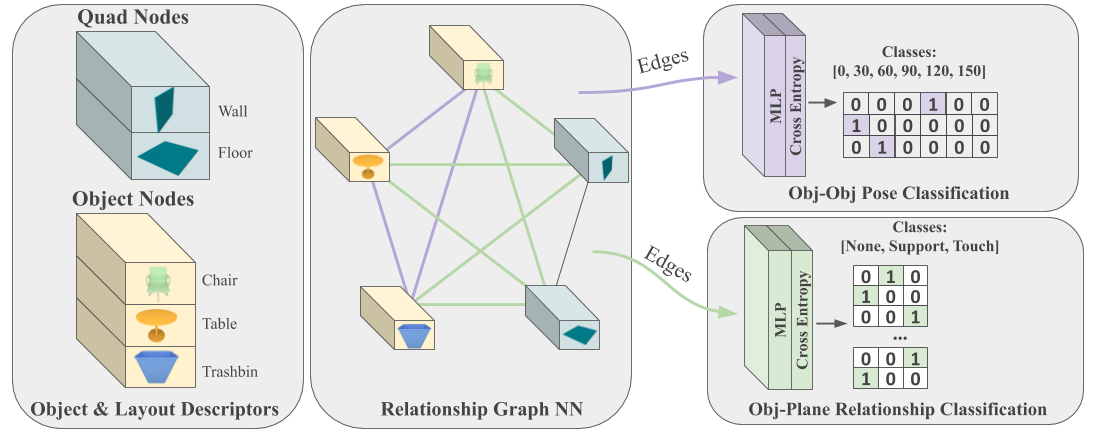}
    \end{center}
   \caption{Object and layout relational prediction. 
   We establish a message-passing neural network in order to predict object-object and object-layout relations. 
   The inputs are feature descriptors of detected objects and quads pooled to the same size, and the output is relationship classification between objects and layout elements, as well as pose relations between objects. 
   Note this relational inference is fully differentiable, enabling end-to-end prediction.
   }
\label{fig:architecture_gnn}
\end{figure}

Figure~\ref{fig:architecture_gnn} shows an overview of our message-passing network.
Messages are passed from nodes to edges for a graph $G = (V,E)$, with nodes $v_i\in V$ and edges $e_{j,k}=(v_j,v_k)\in E$.
We define the message passing similar to \cite{gilmer2017neural,kipf2018neural,dai2019scan2mesh,Wang_2019}:
\begin{equation*}
v\rightarrow e: \mathbf{h}_{i,j}^{t+1} = f_e(\textrm{concat}(\mathbf{h}_i^t, \mathbf{h}_j^t - \mathbf{h}_i^t))
\end{equation*}
where $\mathbf{h}_i^t$ is the feature corresponding to vertex $v_i$ at message passing step $t$, $\mathbf{h}_{i,j}^t$ is the feature corresponding edge $e_{i,j}$ at step $t$, and $f_e$ represents an MLP.
That is, edges features are computed as the concatenation of its constituent vertices.

We then take these output edge features from the message passing and perform a classification of various relationships using a cross entropy loss.
We describe the relationships as follows, which we chose as they do not require extra manual annotation effort given existing ground truth CAD alignments and scene layout; see Section~\ref{sec:extraction_relationship} for more detail regarding extraction of ground truth object and layout relationships.
For object-layout relationships, we formulate a 3-class classification task for support relations, predicting \emph{horizontal support}, \emph{vertical support}, or no support. Only one relationship per object-layout pair is allowed.
For object-object relationships, we predict the angular difference between the front-facing vectors of the respective objects, in order to recognize common relative arrangements of objects (e.g., chairs often face tables). This is trained with a 6-class cross entropy loss where the angular deviation up to $180\circ$ is discretized into 6 bins.

Here, the relationship prediction adds a proxy loss to the model in Figure~\ref{fig:architecture} which inter-correlates object and layout alignments, implicitly guiding the CAD model alignment and layout quad estimation to become more globally consistent.

\section{Object+Layout Dataset}
To train and evaluate our method, we introduce a new dataset of \NUMSCANNETLAYOUT{} CAD layout annotations to the real-world RGB-D scans of the ScanNet dataset~\cite{dai2017scannet}.
These layout annotations, in addition to the CAD annotations of Scan2CAD~\cite{avetisyan2019scan2cad} to ScanNet scenes, inform our method and evaluation on real-world scan data.

In order to obtain these room layout annotations, we use a semi-automated  annotation process.
We then automatically extract the object-object and object-layout relations.

\subsection{Extraction of Scene Layouts}
We performed a semi-automatic layout annotation for ScanNet scene data.
First, large planar surfaces are detected using RANSAC on the reconstructed scans.
We then employ a manual refinement step to modify potential errors in the automatic extraction. The surface extraction is preceded by a semantic instance segmentation to obtain wall, floor, ceiling, window, door, etc. instances. RANSAC is then applied to extract 3D planes from each instance. Planes that fall below a threshold will be merged or connected. All planes are projected onto the floor plane and through a set of various heuristics the most plausible intersection points are selected to ultimately become corner points for the final layout. The room height is either estimated by the maximum height of the detected wall instances or is spanned by the ceiling.

Following the proposals given by RANSAC, we then manually verified which proposals were plausible.
This step is relatively quick ($\approx 2$min per scene) and indicated that the RANSAC produced \NUMSCANNETLAYOUT{} plausible initial layouts.
These layouts were then refined through a manual annotation process.
We developed a Blender\footnote{https://www.blender.org}-based tool was introduced for the layout refinement, allowing annotators to edit/merge/delete corner junctions as well as add or modify edges and planes. 
All automatically generated layouts were verified and refined by two student annotators ($\approx 15min$ per scene).
From the output planes, we then estimate their quad corners and edges as intersections between the planes. 
An illustration of layouts annotation samples on ScanNet can be found in the supplemental.

\subsection{Extraction of Object and Layout Relationships}
\label{sec:extraction_relationship}
To support learning global scene relationships, we extract object and layout relations to supervised our message-passing approach to learning relationships.
We opt to learn relations which can be automatically extracted from given CAD model and layout annotations.

We extract object-object and object-layout relationships. 
For the object-object case, we compute the angular difference between the front-facing vectors of each object where symmetrical properties are ignored; in practice, we compute this on-the-fly during the training process.

Relationships between objects and layout elements are established by support:
\begin{itemize}
  \item A vertical \textbf{support} relationship between a layout element and an object is valid if the bottom side of the bounding box of the object within close proximity to and close to parallel to the layout element.
  \item A horizontal \textbf{touch} relationship is valid if the left, right, front or back side of the bounding box of the object is within close proximity to and close to parallel to the layout element.
\end{itemize}

These relations are  extracted through an exhaustive search. 
That is, each pair of object-layout is checked for vertical support or horizontal touch.
To estimate proximity of objects, we expand the bounding box of the objects by $\tau_p$, and expand the sides of the bounding boxes of the layout elements by $\tau_p$.
We then consider the object and layout element to be in close proximity if their expanded bounding boxes overlap. 
We use $\tau_p=0.2$ meters for all experiments.

\subsection{Synthetic Data}
We additionally evaluate our approach on synthetic data, where CAD object and layout ground truth are provided in the construction of the synthetic 3D scenes.
We use synthetic scenes from the SUNCG dataset~\cite{song2016ssc}. 
SUNCG contains models of indoor building environments including CAD models and room layouts. Layout components are given and hence extraction into planar quads can be performed automatically. 
To generate the input partial scans, we virtually scan the scenes to produce input scans similar to real-world scenarios, following previous approaches to generate synthetic partial scan data~\cite{hou20193dsis}.

Object and layout relational information was extracted following the same procedure for ScanNet data.

\section{Results}
\begin{table}[tp!]
\begin{center}
\resizebox{\textwidth}{!}{%
\begin{tabular}{|l|r r r r r r r r r | r | r|}
\hline
   & bathtub & bookshelf & cabinet & chair & display & other & sofa & table & trashbin         & class avg.       & avg.           \\ \hline
\hline
FPFH (Rusu et al.~\cite{rusu2009fast}) & 0.00 & 1.92 & 0.00 & 10.00 & 0.00 & 5.41 & 2.04 & 1.75 & 2.00 & 2.57 & 4.45 \\ 
SHOT (Tombari et al.~\cite{tombari2010signature}) & 0.00 & 1.43 & 1.16 & 7.08 & 0.59 & 3.57 & 1.47 & 0.44 & 0.75 & 1.83 & 3.14 \\
Li et al.~\cite{li2015database} & 0.85 & 0.95 & 1.17 & 14.08 & 0.59 & 6.25 & 2.95 & 1.32 & 1.50 & 3.30 & 6.03 \\ 
3DMatch (Zeng et al.~\cite{zeng20173dmatch}) & 0.00 & 5.67 & 2.86 & 21.25 & 2.41 & 10.91 & 6.98 & 3.62 & 4.65 & 6.48 & 10.29  \\ 
Scan2CAD (Avetisyan et al.~\cite{avetisyan2019scan2cad})  & 36.20  & 36.40 & 34.00 & 44.26 & 17.89 & \textbf{70.63} & 30.66 & 30.11 & 20.60 & 35.64 & 31.68 \\ End2End (Avetisyan et al.~\cite{avetisyan2019end}) & 38.89 & 41.46 & 51.52 & 73.04 & 26.53 & 26.83 & 76.92 & \textbf{48.15} & 18.18 & 44.61 & 50.72 \\ 
\hline
Ours (dense) & 33.33 & 39.39 & \textbf{58.62} & 70.76 & 28.57 & 33.72 & 50.00 & 34.55 & 23.73 & 41.41 & 51.05 \\
Ours (dense) + obj-obj & 44.44&	\textbf{54.55}&	49.15&	68.05&	37.50&	36.05&	61.11 &	42.01 &	27.12 &	46.66	& 52.97 \\
Ours (dense) + layout & \textbf{54.55} & 47.37 & 38.33 & 71.11 & 32.88 & 28.05 & 62.86 & 37.91 & \textbf{32.26} & 45.04 & 52.06 \\
Ours (dense) full & 39.39 & 42.11 & 48.33 & 74.32 & 42.47 & 36.59 & 62.86 & 36.26 & 30.65 & 45.89 & 54.33 \\
Ours (sparse) & 42.42 & 39.47 & 51.67 & 77.28 & 45.21 & 28.05 & 77.14 & 37.91 & 25.81 & 47.22 & 55.77 \\
Ours (sparse) + obj-obj & 42.42 & 44.74 & 50.00 & 77.53 & 43.84 & 30.49 & 74.29 & 39.56 & \textbf{32.26} & 48.35 & 56.70 \\
Ours (sparse) + layout & 45.45 & 42.11 & 48.33 & 78.27 & 42.47 & 31.71 & 77.14 & 37.36 & 27.42 & 47.81 & 56.29 \\
Ours (sparse) full & 42.42 & 36.84 & 58.33 & \textbf{81.23} & \textbf{50.68} & 40.24 & \textbf{82.86} & 45.60 & \textbf{32.26} & \textbf{52.27} & \textbf{61.24} \\
\hline
\end{tabular}}
\end{center}
\caption{CAD alignment evaluation on ScanNet Scan2CAD data~\cite{dai2017scannet,avetisyan2019scan2cad}. Our final method (last row), incorporating contextual information from both object-object relationships and object-layout relationships, outperforms the baseline by a notable margin of $10.52\%$.}
\label{tab:scannet_eval}
\end{table}

\subsection{CAD Alignment Performance}
\begin{figure}[th!]
     \centering
        \includegraphics[width=0.9\linewidth]{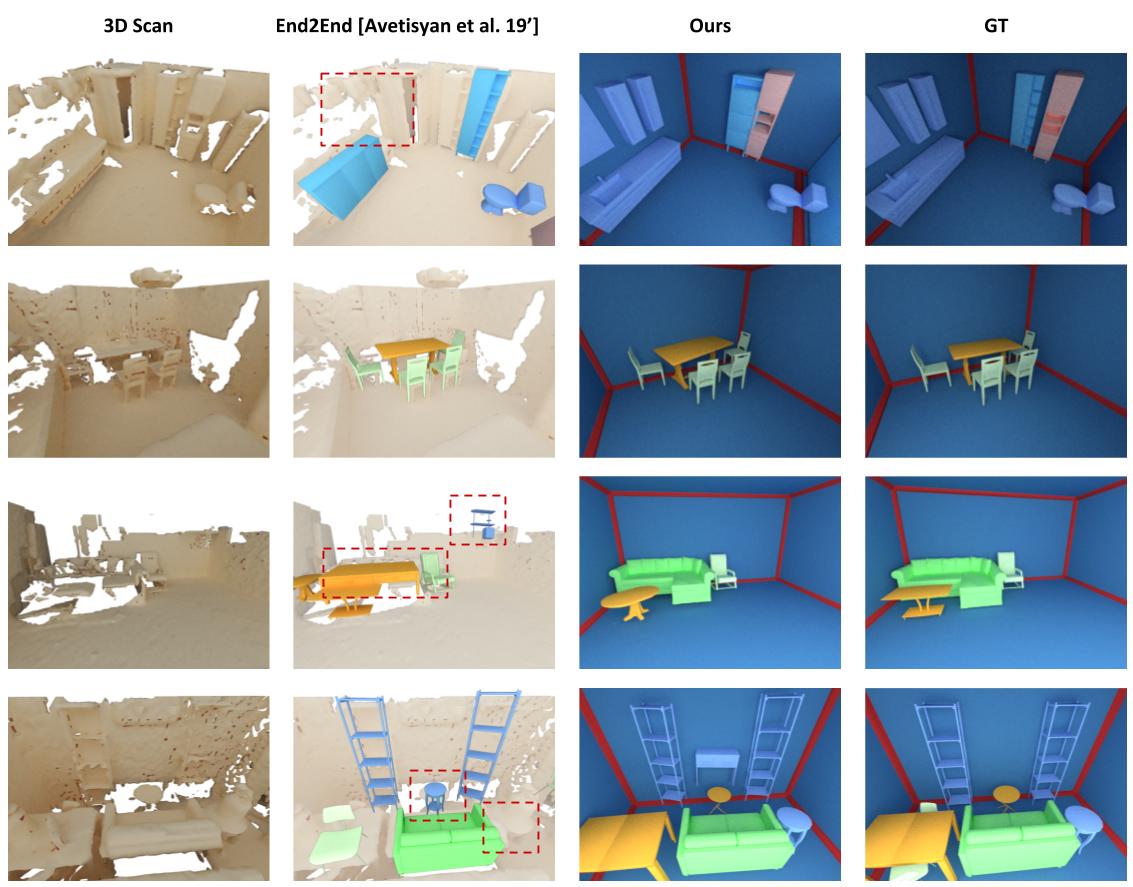}
   \caption{
   Qualitative CAD alignment and layout estimation results on SUNCG~\cite{song2016ssc} scans.
   Our joint estimation approach produces more globally consistent CAD alignments and generates additionally room layout applicable for VR/AR applications. }
   \label{fig:results_suncg}
\end{figure}

\begin{figure}[th!]
     \centering
        \includegraphics[width=0.9\linewidth]{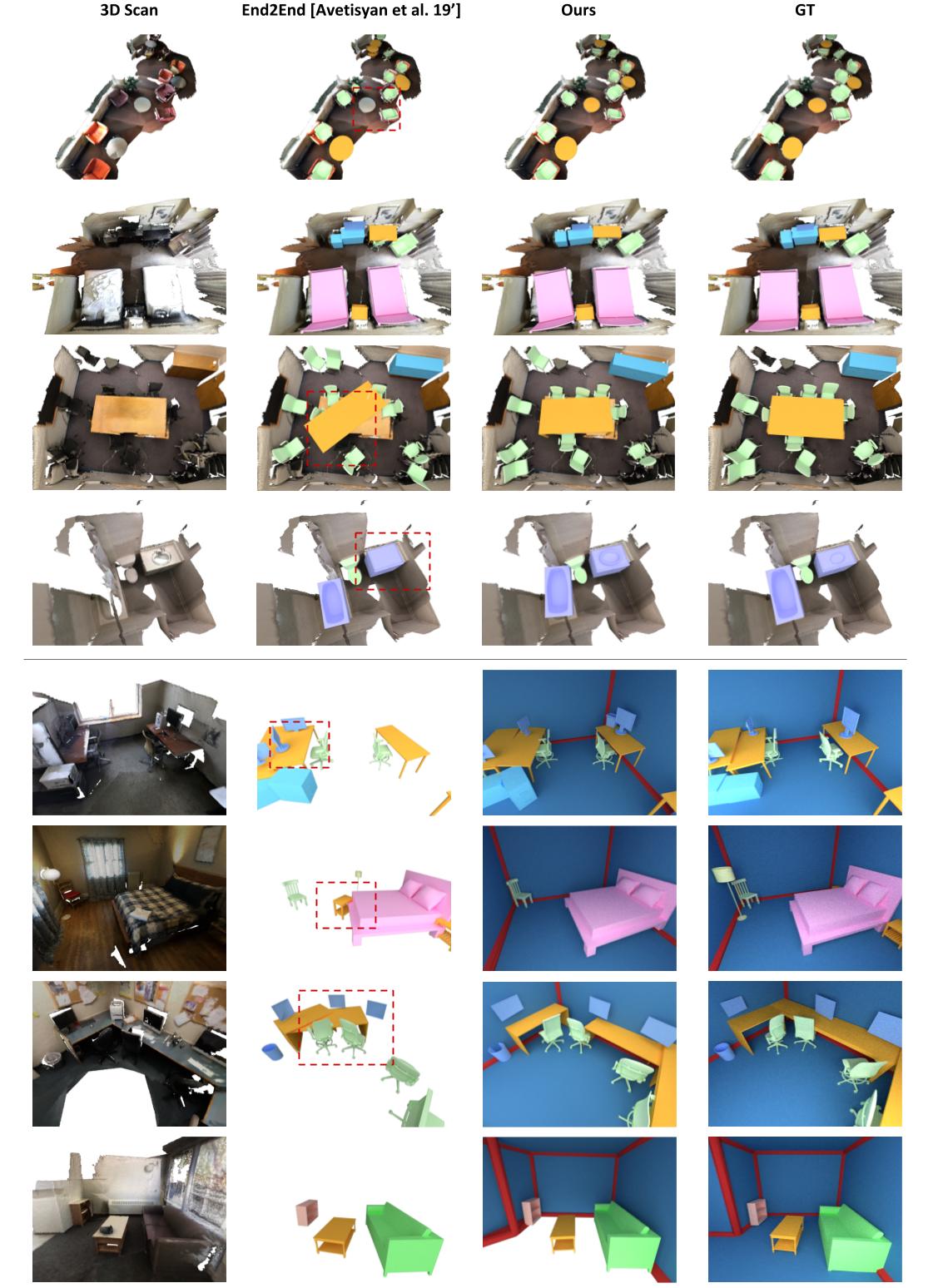}
   \label{fig:results_scannet}
       \caption{Qualitative CAD alignment and layout estimation results on ScanNet~\cite{dai2017scannet} scans (zoomed in views on the bottom). 
       Our approach incorporating object and layout relationships produces globally consistent alignments along with the room layout. 
       }

\end{figure}

We evaluate our method on synthetic SUNCG~\cite{song2016ssc} scans as well as real-world ScanNet~\cite{dai2017scannet} scans in Tables~\ref{tab:suncg_eval} and \ref{tab:scannet_eval}, respectively.
We follow the CAD alignment evaluation metric proposed by \cite{avetisyan2019scan2cad}, which measures alignment accuracy where an alignment is considered successful if it falls within $20$cm, $20^\circ$, and $20\%$ scale of the ground truth.
On both SUNCG and ScanNet scans we compare to several state-of-the-art handcrafted geometric feature matching approaches~\cite{rusu2009fast,tombari2010signature,li2015database} and learned approaches~\cite{zeng20173dmatch,avetisyan2019scan2cad,avetisyan2019end}.
We additionally show qualitative comparisons in Figures~\ref{fig:results_suncg} and \ref{fig:results_scannet}.
On synthetic scan data we outperform the strongest baseline by $16.58\%$, and improve by $10.52\%$ on real scan data. 
This demonstrates the benefit of leveraging global information regarding object and layout relations in improving object alignments.

We also perform an ablation study on the various design choices and impact of relation information.
We evaluate a dense convolutional backbone for our network architecture (\textit{dense}) in contrast to our final sparse convolutional backbone leveraging the sparse convolutions proposed by \cite{choy20194d}.
We additionally show that the object-to-object relational inference (\textit{obj-obj}) as well as layout estimation (\textit{layout}) improve upon no relational inference, and our full method incorporating both object and layout relational inference, the most contextual information, yields the best performance. 

\subsection{Layout Prediction}
\begin{figure}[th!]
     \centering
        \includegraphics[width=0.9\linewidth]{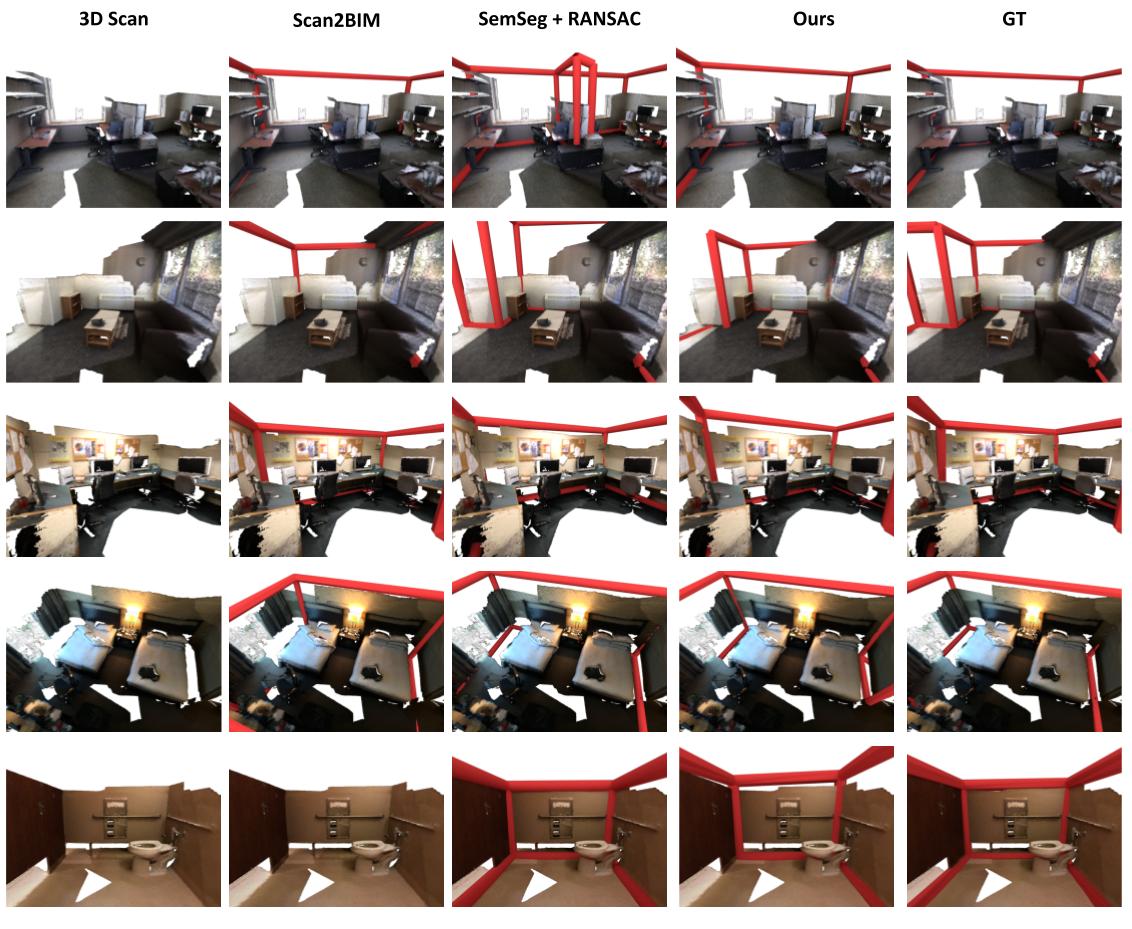}
   \label{fig:results_layout_scannet}
       \caption{Qualitative comparison of our layout estimation on the ScanNet dataset~\cite{dai2017scannet}. 
       Layout elements are highlighted with their wireframes.
       Our method provides a very lightweight, learned approach ($\approx 1M$ trainable parameters) for layout estimation. }

\end{figure}

\begin{figure}[th!]
     \centering
        \includegraphics[width=0.9\linewidth]{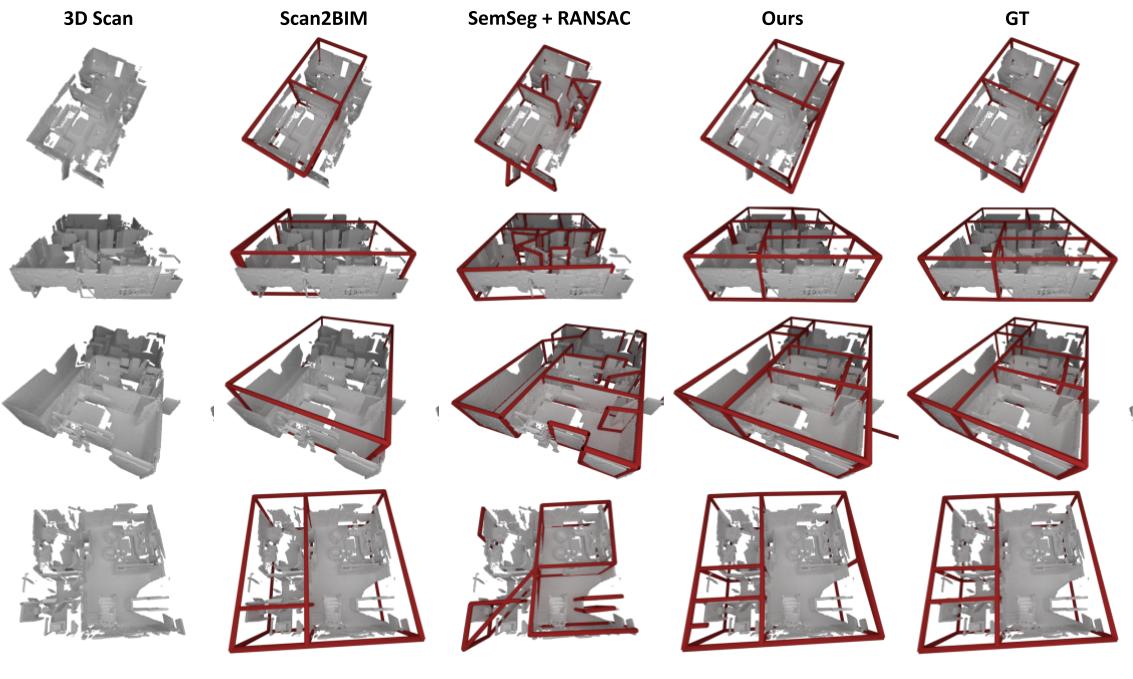}
   \label{fig:results_layout_suncg}
       \caption{Layout estimation on SUNCG~\cite{song2016ssc} scans. 
       Layout elements are highlighted with their wireframes. Our method excels with its simplicity, especially  for very large and complex scenes where heuristics to determine intersections tend to struggle.}

\end{figure}

We evaluate the performance of our layout estimation based on its intermediate corner and edge prediction results as well as the final layout quad predictions.
In Table~\ref{tab:results_layout}, we evaluate the precision and recall of each the corner, edge, and layout quad predictions.
Corners are considered as successfully detected if the predicted corner is within a radius of $40cm$ from the ground truth corner. 
Edges are considered as correctly predicted if they connect the same corners as the ground truth edges. 
Similarly, correctly predicted quads are spanned by the same 4 corners as the associated ground truth quad.
We aim to achieve a high recall for corners and edges due to our hierarchical prediction.
We achieve robust results on both datasets, although ScanNet is notably more difficult as many scenes can miss views of entire layout components (e.g., missing ceilings).

\begin{table}[tp!]
\begin{center}
\resizebox{\textwidth}{!}{%
\begin{tabular}{|l|r r r r r r r r r | r | r|}
\hline
   & bed &	cabinet &	chair &	desk &	dresser &	other &	shelves	& sofa	& table           & class avg.       & avg.           \\ \hline
\hline
SHOT (Tombari et al.~\cite{tombari2010signature}) & 13.43 & 3.23 & 10.18 & 2.78 & 0.00 & 0.00 & 1.75 & 3.61 & 11.93 & 5.21 & 6.30\\
FPFH (Rusu et al.~\cite{rusu2009fast}) & 38.81 & 3.23 & 7.64 & 11.11 & 3.85 & 13.21 & 0.00 & 21.69 & 11.93 & 12.39 & 9.94  \\ 
Scan2CAD (Avetisyan et al.~\cite{avetisyan2019scan2cad})  & 52.24 & 17.97 & 36.00 & 30.56 & 3.85 & 20.75 & 7.89 & 40.96 & 43.12 & 28.15 & 29.23 \\ 
End2End (Avetisyan et al.~\cite{avetisyan2019end}) & 71.64 & 32.72 & 48.73 & 27.78 & 38.46 & 37.74 & 14.04 & 67.47 & 45.87 & 42.72 & 41.83 \\ 
\hline
Ours (dense) & 63.89 & 35.16 & 56.82 & 39.02 & 30.00 & 38.85 & 29.17 & \textbf{76.67} & 31.03 & 44.51 & 44.48 \\
Ours (dense) + obj-obj & 77.78 & 36.26 & 53.03 & 41.46 & 40.00 & 47.48 & 20.83 & \textbf{76.67} & 25.86 & 46.60 & 46.41 \\
Ours (dense) + layout & 75.00 & 37.04 & 60.68 & 37.14 & 38.89 & 45.53 & \textbf{33.33} & 72.41 & 32.08 & 48.01 & 48.33 \\
Ours (dense) full & \textbf{81.25} & 40.00 & 51.92 & 45.45 & 41.18 & 49.17 & 31.58 & 75.86 & \textbf{46.00} & 51.38 & 50.41 \\
Ours (sparse) & 54.29 & 42.55 & 66.67 & 48.57 & \textbf{44.44} & 57.60 & 27.27 & 57.89 & 36.84 & 48.46 & 52.31 \\
Ours (sparse) + obj-obj & 74.29 & 40.43 & 70.09 & \textbf{65.71} & 27.78 & \textbf{60.80} & 27.27 & 55.26 & 38.60 & 51.14 & 55.27 \\
Ours (sparse) + layout & 65.71 & 42.55 & \textbf{77.78} & 54.29 & 38.89 & \textbf{60.80} & 22.73 & 57.89 & 45.61 & 51.81 & 57.12 \\
Ours (sparse) full & 71.43 & \textbf{43.62} & \textbf{77.78} & 54.29 & 38.89 & \textbf{60.80} & 22.73 & 68.42 & 45.61 & \textbf{53.73} & \textbf{58.41} \\
\hline
\end{tabular}}
\end{center}
\caption{CAD alignment accuracy on SUNCG~\cite{song2016ssc} scans. Our final method (last row) goes beyond considering only objects and jointly estimates room layout and  object and layout relationships, resulting in significantly improved performance.}
\label{tab:suncg_eval}
\end{table}

\begin{table}[tp!]
\begin{center}
\footnotesize
\begin{tabular}{|l |l| r | r | r|}
\hline
 Dataset & Metric &corners pred.        & edges pred.          & planes pred.  \\ \hline
\hline
\multirow{2}{*}{SUNCG} & Precision & 80.02 & 78.89 & 64.61  \\ 
& Recall & 84.31 & 91.38 & 96.11  \\ 
\hline
\multirow{2}{*}{ScanNet} & Precision & 58.42 & 64.46 & 58.11  \\ 
& Recall & 70.95 & 74.80 & 80.92  \\ 
\hline
\end{tabular}
\end{center}
\caption{Evaluation of our layout prediction components on SUNCG~\cite{song2016ssc} (top) and ScanNet~\cite{dai2017scannet} (bottom).}
\label{tab:results_layout}
\end{table}
\section{Limitations}
While the focus of this work was to show improved scene understanding through joint prediction of objects \textbf{and} layouts, we believe there is potential for further achievements. For instance, our layout prediction method is bound to predict quad planes only and hence more sophisticated methods could be used for more accurate layout estimation. Also, we used a very lightweight graph neural network for message passing. One could use a more sophisticated method for more accurate relationship prediction and a richer set of relationships that may contain functionality relationships, spatial relationships or room semantic relationships (for instance a bathtub and a shower curtain could be grouped into the same semantic class).

\section{Conclusion}
In this work we formulated a method to digitize 3D scans that goes beyond the focus of objects in the scene. We propose a novel method that estimates the layout of the scene by sequentially predicting corners, then edges and finally quads in a fully differentiable way. The estimated layout is used in conjunction with an object detector to predict contact relationships between objects and the layout and ultimately to predict a CAD arrangement of the scene. We can show that objects and the surrounding (scene layout) go hand in hand and are a crucial factor towards full scene digitization and scene understanding. Objects in the scene are often not arbitrarily arranged, for instance often cabinets are leaned at walls or a table is surrounded by chairs in a dining room, hence we leverage the inherent coupling between objects and layout structure in the learning process. Our approach improves global CAD alignment accuracy by learning those patterns on both real and synthetic scans. 
We hope that we can encourage further research towards this avenue, and see as next immediate steps for future work the necessity of texturing digitized shapes in order to enhance the immersive experience in VR environments. 

\section*{Acknowledgements}
We would like to thank the expert annotators 
Soh Yee Lee, 
Suzana Spasova, and
Danil Bebnev
for their efforts in building our scene layout dataset. 
This work is supported by Occipital, the ERC Starting Grant Scan2CAD (804724), a Google Faculty Award, the ZD.B., a TUM-IAS Rudolf M\"o{\ss}bauer Fellowship, and the German Research Foundation (DFG) Grant \textit{Making Machine Learning on Static and Dynamic 3D Data Practical}.

%
%
\bibliographystyle{splncs04}
\bibliography{egbib}

\newpage
\begin{appendix}

\section{Dataset}
\begin{figure}[H]
     \centering
        \includegraphics[width=0.9\linewidth]{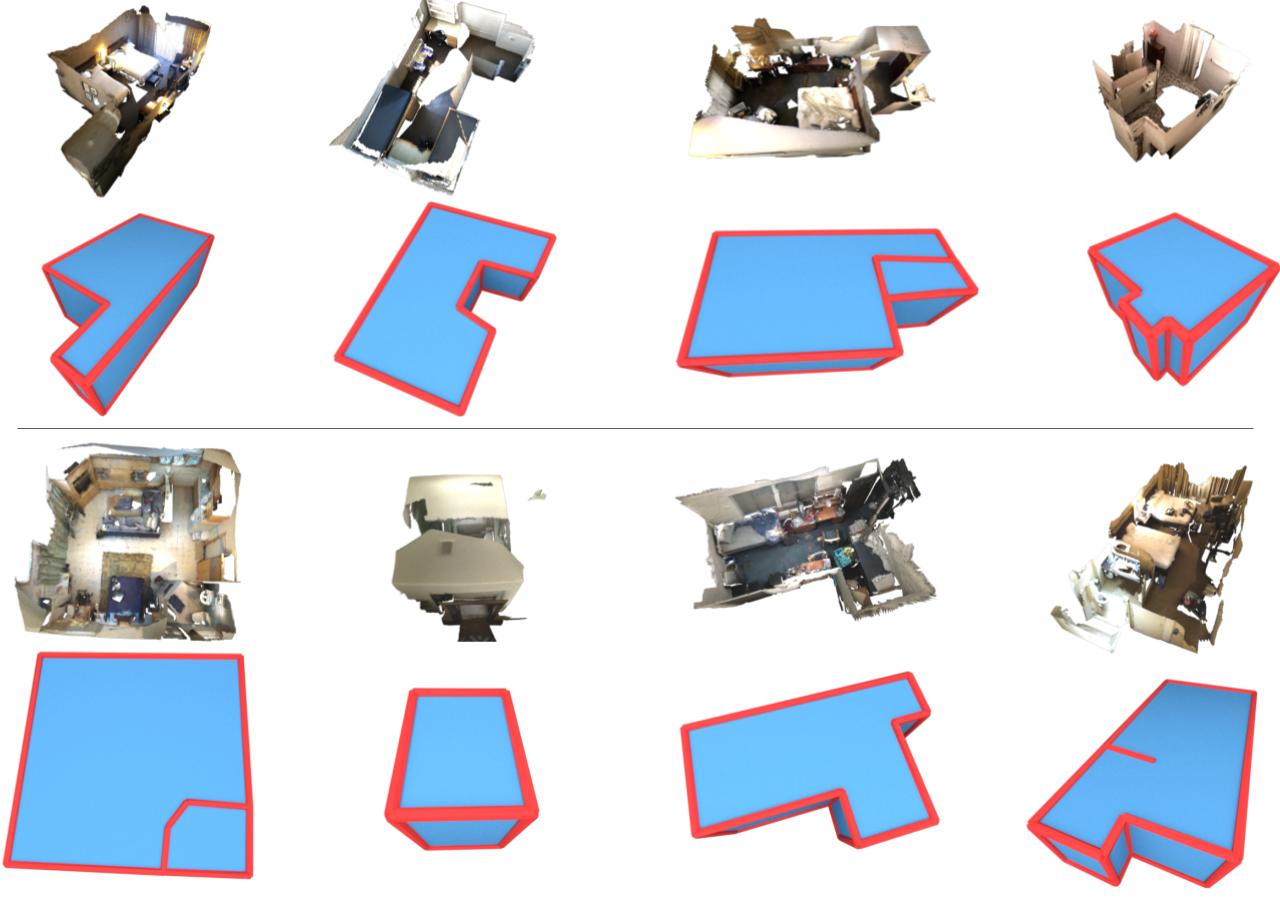}
       \caption{Samples of manually annotated layouts on ScanNet~\cite{dai2017scannet}. Annotations include wireframes and room-level CAD shells of the scenes representing walls, floors, and ceilings.}
          \label{fig:layout_scannet}
\end{figure}

\begin{figure}[H]
     \centering
        \includegraphics[width=0.9\linewidth]{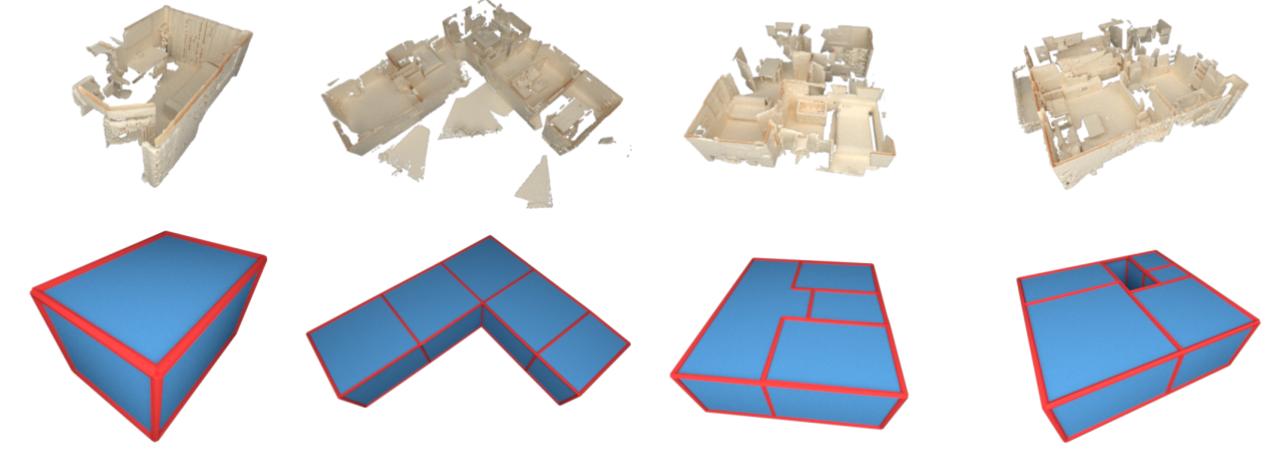}
       \caption{Samples of automatically parsed layouts from SUNCG~\cite{song2016ssc}. Layouts include wireframes and room-level CAD shells.}
          \label{fig:layout_suncg}
\end{figure}

\begin{figure}[H]
\begin{center}
   \includegraphics[width=0.8\linewidth]{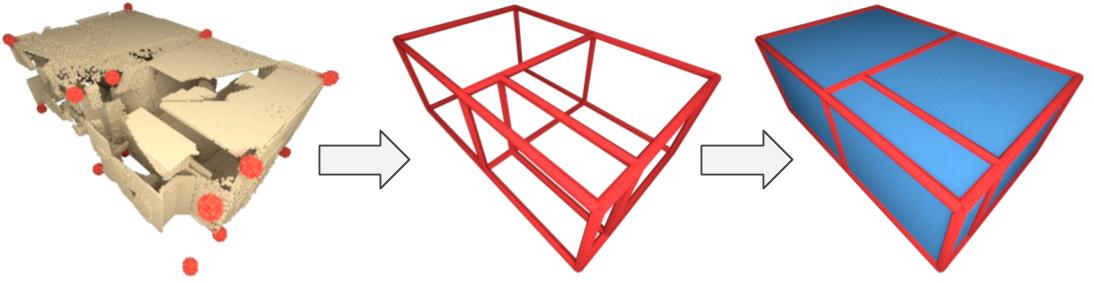}
    \end{center}
   \caption{Sample targets of the layout estimation. The pipeline starts with a corner point estimation (left). Then, valid edges are estimated from the detected corners producing a wireframe (middle). Finally, valid layout quads are predicted from the edge candidates.}
\label{fig:corners_heatmap}
\end{figure}

In this supplemental document, we provide additional details for our layout dataset that is used for training. 
Figure \ref{fig:layout_scannet} shows an illustration of the real-world layout annotations comprising of more than 1000 individual ScanNet~\cite{dai2017scannet} scenes. 
In addition to layouts from real-world scans, we also extract layouts from the synthetic SUNCG dataset~\cite{song2016ssc}; see Figure \ref{fig:layout_suncg}.
From these dataset annotations, we create ground truth targets. as visualized in Figure~\ref{fig:corners_heatmap}. for our hierarchical layout estimation training.

\begin{figure}[H]
\begin{center}
   \includegraphics[width=0.5\linewidth]{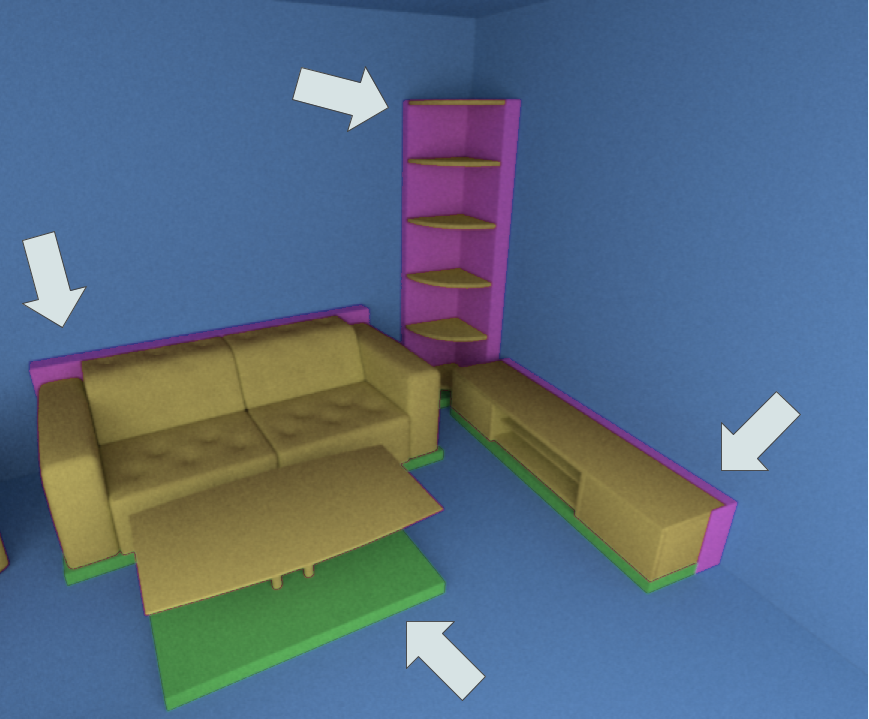}
    \end{center}
   \caption{Groundtruth sample from the dataset. Green tile correspond to \textit{vertical support} relationship and pink tile correspond to \textit{horizontal touch} relationship.}
\label{fig:relationships}
\end{figure}

In Figure \ref{fig:relationships}, we illustrate visually the different kinds of object-to-layout relationship classes. Note that an object can have relationships with multiple layout elements.

\end{appendix}

\end{document}